\PassOptionsToPackage{table}{xcolor}

\documentclass{article}

\usepackage{microtype}
\usepackage{graphicx}
\usepackage{booktabs} 
\usepackage{subcaption}
\usepackage{hyperref}

\usepackage[accepted]{icml2025}

\usepackage{amsmath}
\usepackage{amssymb}
\usepackage{mathtools}
\usepackage{amsthm}
\usepackage[table]{xcolor} 
\usepackage{booktabs} 
\usepackage{siunitx} 
\usepackage{indentfirst}
\usepackage{multirow}
\usepackage{makecell}
\usepackage[capitalize,noabbrev]{cleveref}

\theoremstyle{plain}

\theoremstyle{definition}

\theoremstyle{remark}

\usepackage[textsize=tiny]{todonotes}

\icmltitlerunning{QuarterMap}

\begin{document}

\twocolumn[
\icmltitle{QuarterMap: Efficient Post-Training Token Pruning \\ for Visual State Space Models}



\icmlsetsymbol{equal}{*}

\begin{icmlauthorlist}
\icmlauthor{Tien-Yu Chi}{nycu,canva}
\icmlauthor{Hung-Yueh Chiang}{uta}
\icmlauthor{Diana Marculescu}{uta}
\icmlauthor{Kai-Chiang Wu}{nycu}
\end{icmlauthorlist}

\icmlaffiliation{nycu}{National Yang Ming Chiao Tung University}
\icmlaffiliation{uta}{The University of Texas at Austin}
\icmlaffiliation{canva}{Canva}

\icmlcorrespondingauthor{Tien-Yu Chi}{b03902059@ntu.edu.tw}
\icmlkeywords{State Space Model, Mamba, Efficiency, ICML}

\vskip 0.3in
]



\printAffiliationsAndNotice{}  

\begin{abstract}
State space models (SSMs) reduce the quadratic complexity of transformers by leveraging linear recurrence. Recently, VMamba has emerged as a strong SSM-based vision backbone, yet remains bottlenecked by spatial redundancy in its four-directional scan. We propose QuarterMap, a post-training activation pruning method that removes redundant spatial activations before scanning and restores dimensions via nearest-neighbor upsampling. Our method improves throughput without retraining. On ImageNet-1K, QuarterMap achieves up to 11\% speedup on VMamba with less than 0.9\% accuracy drop, and yields similar gains on ADE20K segmentation. Beyond VMamba, we validate QuarterMap on MedMamba, a domain-specific model that shares the same four-directional scanning structure, where it consistently improves throughput while preserving accuracy across multiple medical imaging tasks. Compared to token merging methods like ToMe, QuarterMap is tailored for SSMs and avoids costly merge-unmerge operations. Our method offers a plug-and-play tool for deployment-time efficiency without compromising transferability.
\end{abstract}
\section{Introduction}
\label{sec:intro}

Advancements in computer vision have been significantly driven by deep learning and the availability of large-scale datasets. Convolutional Neural Networks (CNNs) have served as the basis for tasks such as image classification \cite{AlexNet2012, vgg, he2016deep} and object detection \cite{girshick2014rich, girshick2015fast, redmon2016you}. However, CNNs exhibit limitations in capturing long-range dependencies. Vision Transformers (ViTs) \cite{ViT2021, Swin2021, DeiT2021}, with their self-attention mechanisms, effectively overcome these limitations, but incur high computational costs due to their quadratic complexity. To alleviate these computational demands, recent research has focused on reducing the complexity of ViTs \cite{wang2020linformer, beltagy2020longformer, Swin2021, liu2022swin, liu2023efficientvit}, applied model compression techniques \cite{liu2021post, lin2021fq, zhu2021vision, DeiT2021, lin2023supervised}, and investigated alternative architectures such as RWKV \cite{rmkvpeng2023rwkv} and State Space Models (SSMs) \cite{s4gu2021, h3fu2022hungry, gu2023mamba}.

\begin{figure}
    \centering
    \includegraphics[width=\columnwidth]{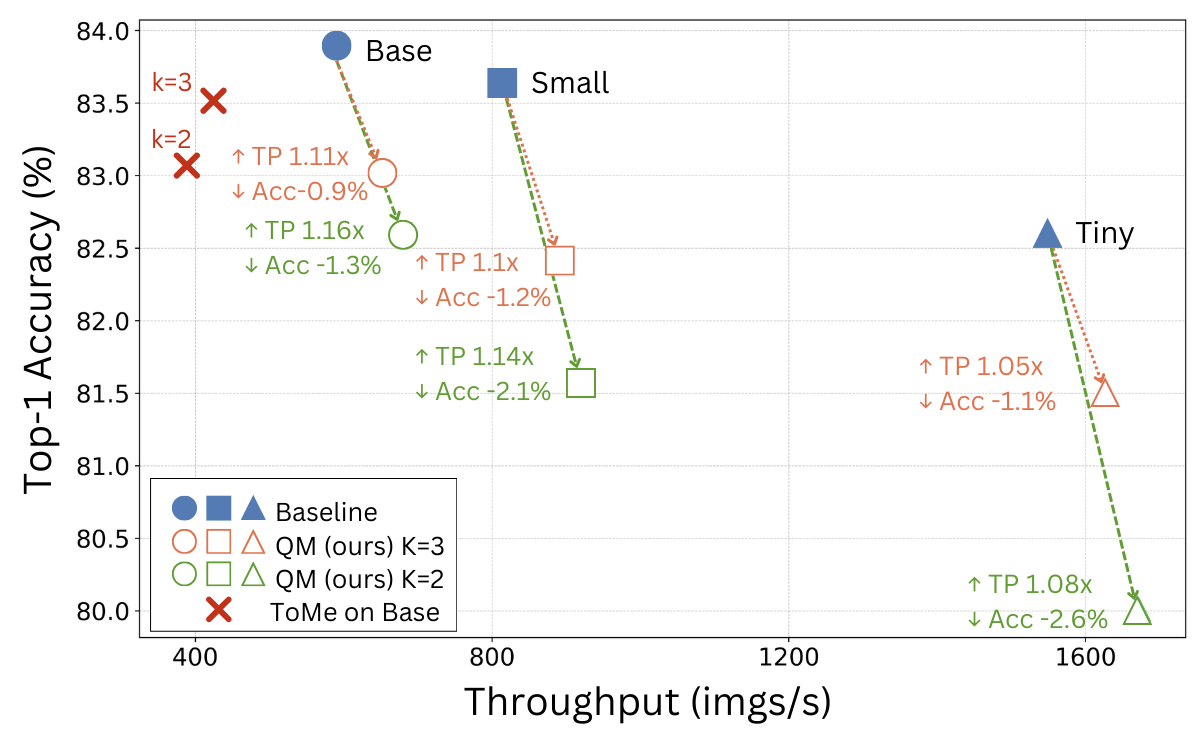}
    \caption{The accuracy-throughput trade-off when applying QuarterMap and Token Merging (ToMe). Our method demonstrates that QuarterMap not only increases throughput but also maintains comparable accuracy. In contrast, ToMe experiences a drop in throughput due to the overhead of merge and unmerge operations.}
    \label{fig:1}
\end{figure}







\begin{figure*}[ht]
    \centering
    \includegraphics[width=\textwidth]{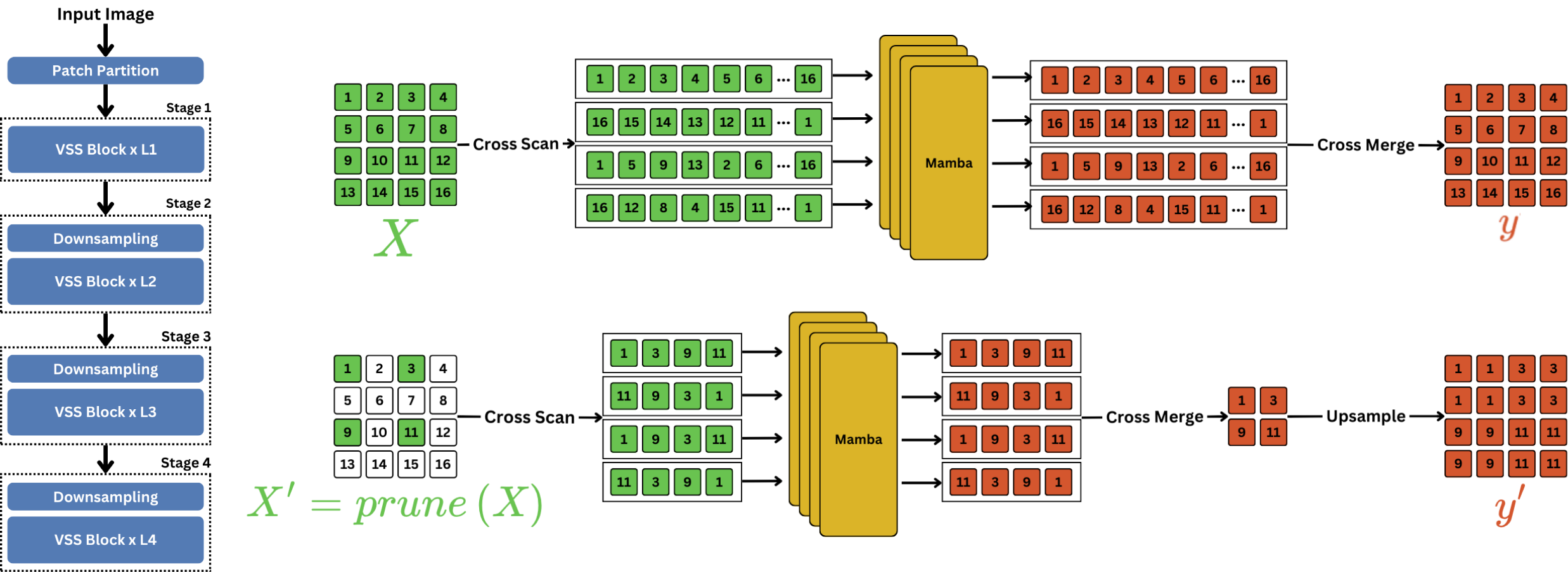}
    \caption{
Illustration of the VMamba model architecture (\textbf{left}) and the proposed QuarterMap applied to the SS2D mechanism (\textbf{right}). The top flow shows the original cross-scan and cross-merge operations, while the bottom applies QuarterMap, pruning activations before scan and restoring spatial dimensions via nearest-neighbor upsampling. This reduces spatial redundancy and improves runtime efficiency without retraining.
    }
    \label{fig:vmamba}
\end{figure*}

State Space Models (SSMs) were initially introduced in the natural language processing (NLP) domain to reduce the computational cost associated with maintaining hidden states during the decoding phase. In contrast, computer vision tasks typically interpret the hidden state as a representation of the entire image's information. Recently, SSM have emerged as efficient alternatives to ViTs in computer vision, demonstrating competitive performance across multiple tasks \cite{liu_vmamba_2024, zhu2024vision, Yang_2024_BMVC, li2024videomamba, teng2024dim}. For instance, VMamba \cite{liu_vmamba_2024} achieves a top-1 accuracy of 82.6\% on the ImageNet-1K benchmark \cite{deng2009imagenet}, outperforming the Swin Transformer \cite{Swin2021} by 1.3\% with comparable FLOPs. However, within VMamba, the kernel responsible for selective scanning, an operation analogous to attention in transformer models, both serving as mechanisms to capture global context, still accounts for 18.3\% of the total kernel execution time, highlighting a notable efficiency bottleneck. To address this, we explored transformer optimization techniques but identified a lack of optimization methods specifically designed for SSMs. We prove that conventional methods, such as token merging \cite{bolya2023tokenmergingvitfaster} widely used in transformers, are suboptimal for VMamba due to computational tradeoffs arising from frequent merging and unmerging operations, as illustrated in \cref{fig:1}. Other recent approaches, such as Top-ViM \cite{zhan2024exploring} and R-MeeTo \cite{shi2025fastervisionmambarebuilt}, introduce token pruning and merging strategies tailored to ViM-based \cite{zhu2024vision} models. However, both rely on retraining to maintain performance, making them less applicable with limited computational resources in post-training deployment scenarios.

Motivated by these challenges, we explore whether a technique similar to token merging could be adapted for activation pruning in SSMs, and more broadly, how efficiency can be further improved in already efficient linear SSMs without \emph{retraining}. We begin by analyzing VMamba's cross-scan and cross-merge mechanism, and the effective receptive field (ERF) in VMamba \cite{liu_vmamba_2024}. Our analysis reveals that the four-directional traversal in VMamba, which introduces substantial spatial redundancy, with some tokens accumulating excessive and potentially unnecessary information. This led us to hypothesize that specialized activation pruning, aligned with the scanning structure, could reduce the latency while preserving accuracy.

To this end, we propose QuarterMap, a training-free activation pruning method specifically designed to improve VMamba’s efficiency by reducing the feature map to one quarter of its original spatial size before scanning. As illustrated in \cref{fig:vmamba}, QuarterMap introduces a lightweight two-stage pipeline: spatial pruning is applied before the cross-scan module, and nearest-neighbor upsampling is used to restore resolution after cross-merge. During pruning, QuarterMap retains every alternate element in both spatial dimensions, leveraging the spatial redundancy inherent in VMamba’s four-way scanning. We adopt nearest-neighbor interpolation under the hypothesis that adjacent spatial positions convey similar information, making it an efficient yet effective reconstruction strategy. This design significantly reduces computation across the cross-scan, selective scan, and cross-merge, all without modifying model weights or requiring retraining.

We evaluate QuarterMap on both image classification and semantic segmentation tasks, showing improved throughput with minimal accuracy degradation. On ImageNet-1K, it achieves up to a \( 1.11 \times \) speedup with less than a 0.9\% drop in top-1 accuracy. Similar trends are observed in semantic segmentation and medical imaging benchmarks. QuarterMap proves particularly effective for VMamba and its variants, as confirmed by comparisons with CNNs, ViTs, and other SSMs like PlainMamba \cite{Yang_2024_BMVC} and ViM \cite{zhu2024vision}. Visualizations of attention maps and effective receptive fields (ERF) show that QuarterMap removes redundant activations while preserving key spatial signals. Comprehensive ablation studies further explore pruning intervals, layer selection strategies, and upsampling methods, providing practical guidance for training-free deployment in the real world.
\section{Method}
\label{sec:method}

We introduce QuarterMap, a post-training pruning function specifically designed to enhance the efficiency of VMamba by reducing spatial redundancy in activation feature maps, as illustrated in \cref{fig:vmamba}. Formally, we define QuarterMap as a function \( T \) that operates on an input activation map \( x \) by selectively retaining spatial information. Given an activation map \( x \in \mathbb{R}^{H \times W \times D} \), where \( H \) and \( W \) represent the spatial dimensions and \( D \) denotes the channel dimension, QuarterMap operates through the following stages:

\noindent \textbf{Block Selection} QuarterMap applies pruning selectively to specific blocks within the VMamba architecture, determined by a block selection interval \(k\). This high-level strategy governs how frequently pruning is applied, balancing computational efficiency and accuracy. Applying QuarterMap to every three blocks (\emph{i.e.,} k=3), excluding the first layer, yields the best accurarcy-latency trade-off (\emph{ref.} \cref{appendix:ablation_study}.) The early layers are critical for encoding fundamental features, whereas the deeper layers are more resilient to pruning, making them suitable candidates for optimization.

\begin{figure}[ht]
    \centering
    \includegraphics[width=0.5\columnwidth]{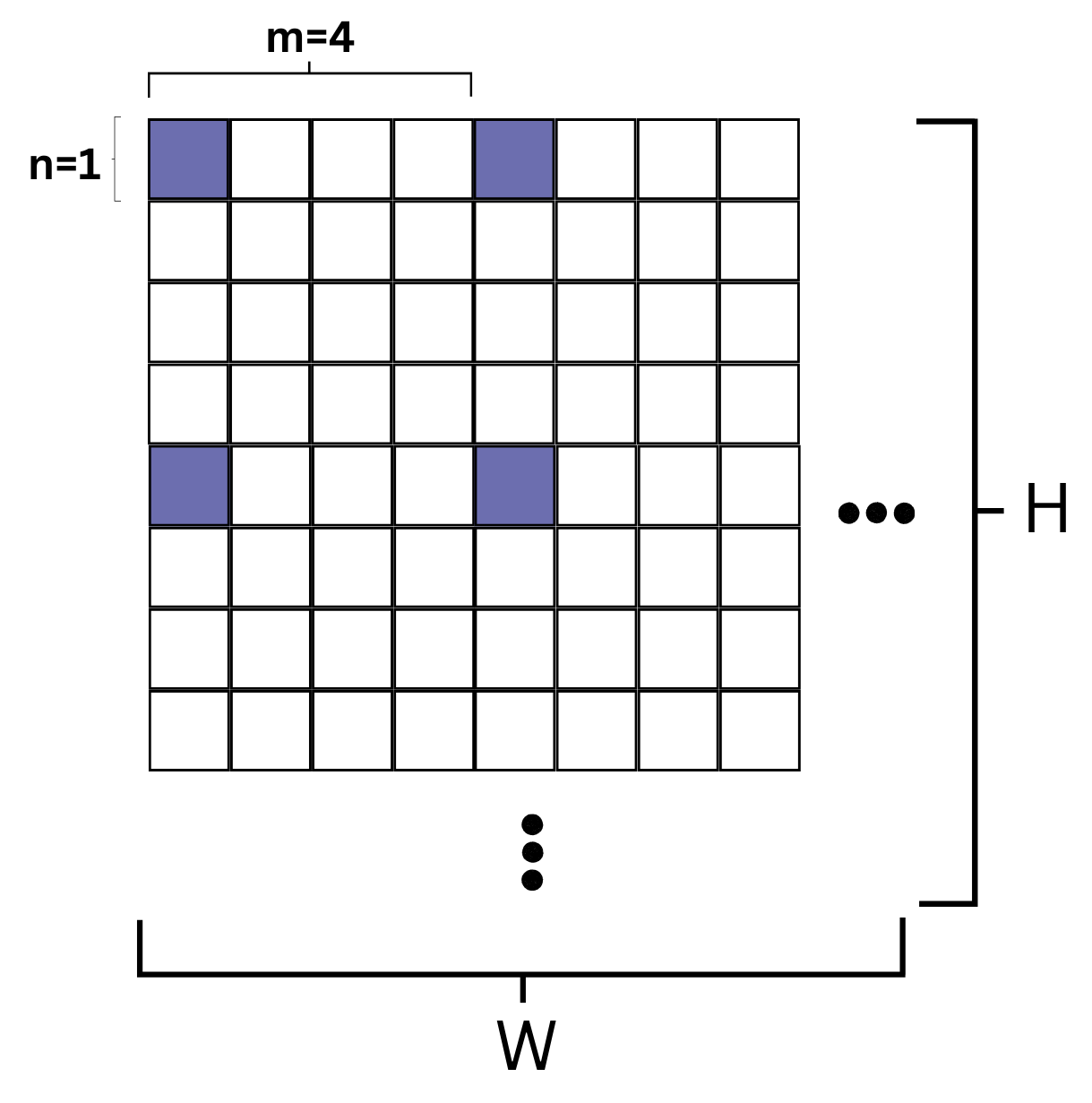}
    \caption{Illustration of the pruning stage, where \(m\) represents the interval size and \(n\) indicates the number of elements retained in the spatial dimensions.}
    \label{fig:pruning_m_n}
\end{figure}

\noindent \textbf{Pruning Stage} Within each selected block, QuarterMap performs a downsampling operation on the spatial dimensions of \( x \) before cross-scan. For a specified interval \( m \), the function \( T \) retains every \( n \) elements in both the \( H \) and \( W \) dimensions, as depicted in \cref{fig:pruning_m_n}, resulting in a pruned activation map \( x' = T(x) \in \mathbb{R}^{\lceil H*n/m \rceil \times \lceil W*n/m \rceil \times D} \). This process leverages the VMamba cross-scan mechanism, which aggregates information from four directions, combined with the SSM recurrence function. Together, these ensure that each element of the feature map \( x_{i,j} \) incorporates information from neighboring spatial positions, improving pruning effectiveness while minimizing the loss of accuracy. We find that the setting \( m = 2 \) and \(n=1\) achieves the optimal trade-off between computational efficiency and accuracy. As the pruning stage is applied prior to cross-scan, the computational savings primarily stem from the reduced size of the input \(x\) for both the cross-scan and Mamba operations. Notably, in Mamba, this reduction leads to savings in the input length for the SSM  (\cref{eq:3}) as well as in the linear computation through the selective mechanism (\cref{eq:6}).

\noindent \textbf{Upsampling Stage} After processing through cross-scan, selective scan, and cross-merge, QuarterMap restores the spatial dimensions of the activation map using an upsampling function \( U \) applied to \( y \), the cross-merge output. Nearest-neighbor interpolation is used to reconstruct the original spatial dimensions, producing an output \( y' = U(y) \in \mathbb{R}^{H \times W \times D} \). This approach aligns with the assumption that adjacent spatial elements contain similar information, enabling QuarterMap to balance computational efficiency with minimal accuracy degradation.

\section{Experiments}
\label{sec:experiments}

\begin{table*}[ht]
\centering
\small

\caption{Performance of VMamba models on ImageNet-1K with and without QuarterMap (QM). QM improves throughput with minimal accuracy drop, while incurring lower overhead than ToMe.}
\begin{tabular}{lccccccc}
\hline
\textbf{Model} & \textbf{Method}       & \textbf{K} & \textbf{Acc@1 (\%)} & \textbf{Throughput} & \textbf{Speedup} & \textbf{Scan Kernel} & \textbf{Add. overheadd} \\
\hline
VMamba-B       & Baseline              & -          & 83.88               & 590            & 1×               & 1.4ms	  & - \\
               & ToMe                  & 3          & 83.52 (-0.36)       & 424            & 0.72×            & \multirow{2}{*}{0.7ms} & \multirow{2}{*}{9.7ms}\\
               & ToMe                  & 2          & 83.07 (-0.81)       & 389            & 0.66×            &  &\\
               & QM (ours)             & 3          & 83.02 (-0.86)       & 654            & 1.11×            & \multirow{2}{*}{0.6ms} & \multirow{2}{*}{0.2ms}\\
               & QM (ours)             & 2          & 82.58 (-1.30)       & 682            & 1.16×            &  &\\
\hline
\end{tabular}
\label{tab:quartermap_classification_base_perf}
\end{table*}

\subsection{Classification and Segmentation}
\label{subsec:classification_and_segmentation}

The results in \cref{tab:quartermap_classification_base_perf} show that applying QuarterMap to the \textit{base} VMamba model yields a slight accuracy drop of 0.86\% when \(k=3\), while improving throughput by \( 1.11 \times \). This gain primarily stems from reduced sequence length within the selective scan mechanism, with minimal overhead introduced by our pruning and upsampling stages. Profiling reveals that QuarterMap reduces the scan kernel time from 1.4 to 0.6 \(ms\), and introduces only 0.2 \(ms\) of additional overhead. In contrast, ToMe incurs significantly higher additional overhead (9.7 \(ms\)), despite a similar scan kernel runtime (0.7 \(ms\)), due to the cost of merge and unmerge operations. These comparisons highlight the efficiency of QuarterMap in minimizing unnecessary computation while retaining accuracy.  Additional results on \textit{small} and \textit{tiny} VMamba configurations, as well as detailed segmentation metrics, are included in Appendix~\ref{appendix:additional_exp}.

\subsection{QuarterMap on MedMamba for Medical Imaging}
\begin{table}[ht]
\centering
\small
\caption{QuarterMap on MedMamba-T across MedMNIST tasks. TP = Throughput (img/s).}
\rowcolors{2}{gray!10}{white}
\begin{tabular}{lcccc}
\toprule
\textbf{Dataset} & \textbf{Base} & \textbf{QM (Ours)} & \textbf{Base} & \textbf{QM (Ours)} \\
                 & \textbf{Acc.} & \textbf{Acc.} & \textbf{TP} & \textbf{TP} \\
\midrule
BloodMNIST  & 97.72\% & 97.78\% & 854.3 & 1033.7 \\
OrganMNIST  & 81.85\% & 81.90\% & 854.5 & 1033.8 \\
RetinaMNIST & 54.25\% & 54.25\% & 853.6 & 1033.8 \\
PathMNIST   & 29.44\% & 29.44\% & 854.0 & 1033.9 \\
\bottomrule
\end{tabular}
\label{tab:quartermap_medmamba}
\end{table}

To evaluate QuarterMap's applicability beyond VMamba, we apply it to MedMamba-T \cite{yue2024medmambavisionmambamedical}, a domain-specific model built on VMamba’s architecture. We benchmark performance across four MedMNIST classification datasets: BoodMNIST, OrganMNIST, RetinaMNIST, and PathMNIST \cite{medmnistv1, medmnistv2}. As shown in \cref{tab:quartermap_medmamba}, QuarterMap consistently delivers a \(1.21\times\) throughput improvement (from ~854 to ~1034 images/sec) with no degradation in classification accuracy. These results demonstrate QuarterMap’s generalization to VMamba-based models and reinforce its utility for domain-specific deployments where post-training efficiency and accuracy preservation are critical. For a breakdown of class-wise performance on BloodMNIST, see Appendix~\ref{appendix:additional_exp}.
\begin{table}[!htbp]
\centering
\small
\caption{Accuracy comparison of QuarterMap (QM) on different model types for ImageNet-1K classification.}
\rowcolors{2}{gray!10}{white}
\begin{tabular}{llSS}
\toprule
\textbf{Model}   & \textbf{Type}      & \textbf{Baseline} & \textbf{QM (Ours)} \\
\midrule
ConvNeXtv2-B     & Conv              & 84.89             & 45.71              \\
DeiT-B           & Transformer       & 81.80             & 79.50              \\
Swin-B           & Transformer       & 85.17             & 82.91              \\
ViM-B            & Mamba             & 80.40             & 71.00              \\
VMamba-B         & Mamba             & 83.88             & 83.02              \\
\bottomrule
\end{tabular}
\label{tab:quartermap_base_arch}
\end{table}
\begin{figure}[ht]
    \centering
    \includegraphics[width=\columnwidth]{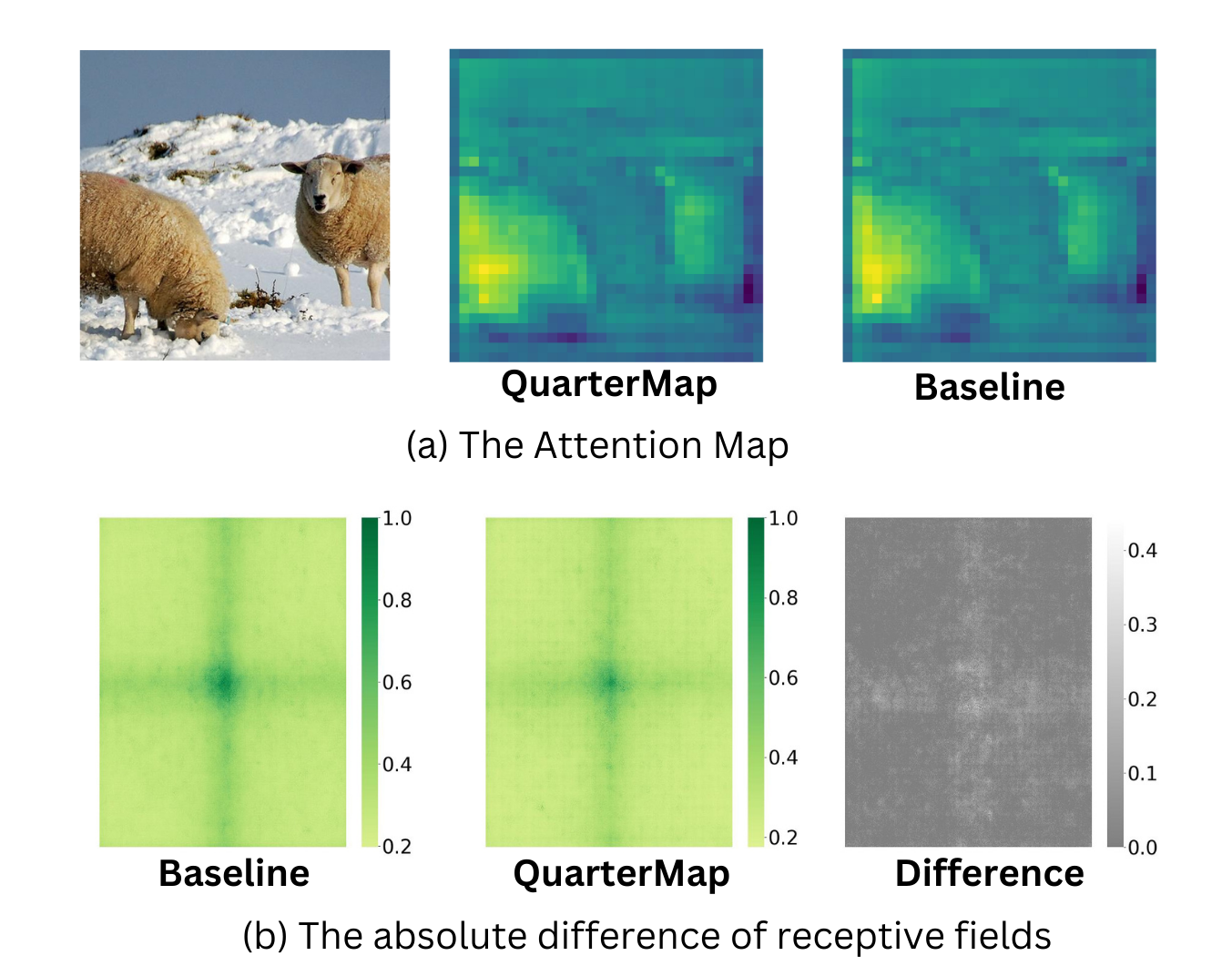}
    \caption{Comparison of (a) attention maps and (b) effective receptive fields before and after applying QuarterMap. The visualizations highlight the differences introduced by QuarterMap, demonstrating its selective pruning of redundant information while preserving the model's essential functionality.}
    \label{fig:attn_map_erf}
\end{figure}
\subsection{QuarterMap on Other Architectures}
We evaluate QuarterMap on CNNs, ViTs, and SSMs to assess whether its design is specifically suited for VMamba. We focus on base variants of ConvNeXtv2 \cite{woo2023convnextv2codesigningscaling}, DeiT \cite{DeiT2021}, Swin Transformer \cite{Swin2021}, and ViM \cite{zhu2024vision}, all pretrained on ImageNet-1K with weights from Hugging Face \cite{wolf2020huggingfacestransformersstateoftheartnatural}. QuarterMap is applied to every third (\emph{i.e.} k=3) blocks after the first two.
In CNNs, pruning disrupts spatial continuity and significantly harms accuracy. ViTs and ViM are more resilient, but still show non-trivial accuracy drops. As shown in \cref{tab:quartermap_base_arch}, our method reduces the latency on VMamba with its four-directional scanning mechanism while maintaining the accuracy.
QuarterMap is less compatible with CNNs and 1D-scanning SSMs like ViM, as these models lack the redundant activation patterns presented in VMamba. See Appendix~\ref{appendix:additional_exp} for results on other variants.

\subsection{Attention Map and Effective Receptive Field (ERF)}
We visualize the attention maps and ERF before and after applying QuarterMap to analyze its spatial impact. Following the formulation in VMamba, we extract the attention map of the \(12^{\text{th}}\) block (\emph{i.e.,} the deepest layer), after several preceding blocks have been pruned. As shown in \cref{fig:attn_map_erf}, the attention patterns remain largely unchanged, indicating that QuarterMap preserves key contextual behavior.
For the ERF visualization, the gray regions represent the activations removed by QuarterMap. These areas often overlap with receptive field hotspots that were spatially redundant. This supports our hypothesis that QuarterMap effectively eliminates redundant information while preserving the functional structure of the model.
\section{Conclusion}
\label{sec:conclusion}

We propose QuarterMap, a post-training pruning method for VMamba that improves runtime efficiency with minimal accuracy loss and requires no retraining. Our experiments show that QuarterMap aligns particularly well with VMamba's four-directional scan and is also compatible with derived applications based on its backbone. Although designed for VMamba, QuarterMap is orthogonal to techniques such as quantization, enabling further efficiency gains. While this study focuses on VMamba, our findings open new directions to understanding and extending pruning strategies in SSMs.

\section*{Acknowledge}

This work was supported in part by the ONR Minerva program, NSF CCF Grant No. 2107085, iMAGiNE - the Intelligent Machine Engineering Consortium at UT Austin, UT Cockrell School of Engineering Doctoral Fellowships

\section*{Impact Statement}

This paper presents work whose goal is to advance the field of 
Machine Learning. There are many potential societal consequences 
of our work. By making ML models more efficient, our work contributes to democratizing access to advanced AI technologies and enabling their wider adoption.
\nocite{langley00}

\bibliography{example_paper}
\bibliographystyle{icml2025}

\newpage
\appendix
\onecolumn
\section{Related Work}
\label{sec:related_work}

\subsection{State Space Models}
State Space Models (SSMs) \cite{hippogu2020, s4gu2021, s5smith2022simplified, h3fu2022hungry, gu2023mamba} have gained notable traction in recent for their efficient scaling properties, offering linear computational complexity with respect to sequence length, which provides an advantage over the quadratic complexity of transformers. This efficiency, combined with their ability to capture global context, has made SSMs an appealing choice for handling long-range dependencies. To further minimize the resource demands of SSMs, S4 \cite{s4gu2021} applied a structured approach using a diagonal matrix configuration enhanced with a low-rank update, which reduced computational overheads. Building on this, subsequent works such as S5 \cite{s5smith2022simplified} and H3 \cite{h3fu2022hungry} introduced innovations including parallel scanning methods and optimized hardware utilization, further advancing the efficiency of SSM-based architectures. With Mamba \cite{gu2023mamba}, the introduction of the S6 block mark a key advancement by incorporating data-dependent parameters, breaking from the Linear Time Invariant (LTI) constraints of earlier models and enabling SSMs to outperform transformers on large-scale datasets.

In the field of vision tasks, S4ND \cite{nguyen2022s4nd} is one of the first models to adapt SSMs for visual data processing, representing it as 1D, 2D, and 3D signals. ViM \cite{zhu2024vision} and VMamba \cite{liu_vmamba_2024} further integrated SSMs into vision backbones, introducing the ViM and VSS blocks, respectively, which employ multiple scanning directions to handle the non-sequential characteristics of image data. This adaptation enables these models to achieve competitive performance with ViTs and CNNs. The success of ViM and VMamba has since inspired a range of Mamba-based methods that address diverse vision tasks, including medical image segmentation \cite{ma2024u, wang2024mamba, xing2024segmamba}, video understanding \cite{li2025videomamba, tang2024vmrnn}, and image generation \cite{hu2024zigma, teng2024dim}. Collectively, these contributions highlight the potential of SSM-based approaches to drive advancements in computer vision applications.

\subsection{Pruning}

Neural network pruning has been a pivotal technique in the evolution of deep learning, aiming to enhance model efficiency by reducing computational and memory demands. Early foundational work \cite{lecun_optimal_1989} introduced the concept of Optimal Brain Damage, which systematically removes less significant weights based on their second-order derivatives, effectively reducing model complexity without substantial loss in performance. This approach lays the groundwork for subsequent pruning methods, including weight pruning and activation pruning.

Weight pruning \cite{han2015deep, frankle2018lottery, li2016pruning, he2018soft} involves eliminating less essential weights from the model, effectively reducing the number of parameters and computational load, making it well-suited for deployment in resource-constrained environments. In contrast, activation pruning \cite{zhuang2018discrimination, yu2018nisp, lin2020hrank, he2017channel} focuses on the intermediate outputs of the network, pruning redundant activations during inference to reduce computational costs without modifying the learned parameters.

The rise of ViTs \cite{dosovitskiy2020image, liu2022swin, DeiT2021} introduced significant computational challenges, leading to the development of token pruning and token merging techniques to shorten sequence lengths in the attention layers. Token pruning \cite{yin2022adavitadaptivetokensefficient, rao2023dynamicspatialsparsificationefficient, rao2021dynamicvit, tang2022patch} selectively removes less informative tokens, reducing the computational burden of self-attention layers while preserving accuracy. Alternatively, token merging \cite{bolya2023token, bolya2023tokenmergingvitfaster} combines similar tokens, effectively decreasing token count and enhancing throughput, thereby preserving ViT performance while reducing the token dimension.

Pruning techniques in the context of State Space Models (SSMs) have limited explored. A pruning-aware hidden state alignment method \cite{zhan2024exploring} is introduced to stabilize the neighborhood of remaining tokens and improve performance. They also propose a token importance evaluation method tailored for SSMs to guide token pruning, though this method was applied only to ViM \cite{zhu2024vision}. R-MeeTo \cite{shi2025fastervisionmambarebuilt} proposes a merged-token re-training strategy that periodically combines similar token pairs in ViM and then fine-tunes the model to restore accuracy. While this achieves good trade-offs, it requires retraining and is not applicable in purely post-training deployment pipelines. EfficientVMamba \cite{pei2024efficientvmambaatrousselectivescan} integrates an atrous-based selective scan approach with efficient skip sampling. This approach introduces a concept similar to pruning, but it is applied before training.
\section{Preliminaries}
\label{appendix:preliminaries}

In this section, we provide an overview of the State Space Model (SSM) \cite{Kalman1960filter} and introduce two recent methods that leverage SSM in innovative ways: the selective state space model (Mamba) \cite{gu2023mamba} and VMamba \cite{liu_vmamba_2024}.

\subsection{State Space Model} 

The State Space Model is a mathematical framework for modeling the evolution of a system over time. It represents the relationship between the system’s state and observations at each time step through a set of equations. The most general form of an SSM is a continuous-time linear dynamical system, as shown in \cref{eq:1}.
\begin{equation} \label{eq:1}
    \begin{gathered}
        h'(t) = A(t)h(t) + B(t)u(t) \\
        y(t) = C(t)h(t) + D(t)u(t)
    \end{gathered}
\end{equation}

Here, \( h(t) \in \mathbb{R}^n \) denotes the state variable at time \( t \in \mathbb{R} \), often referred to as the hidden state in recent machine learning literature. The input is represented by \( u(t) \in \mathbb{R}^m \), and the output by \( y(t) \in \mathbb{R}^p \). The system matrices \( A(t) \in \mathbb{R}^{n \times n} \), \( B(t) \in \mathbb{R}^{n \times m} \), \( C(t) \in \mathbb{R}^{p \times n} \), and \( D(t) \in \mathbb{R}^{p \times m} \) govern the system dynamics at each time step. For simplicity, we consider \( u(t) \) and \( y(t) \) as scalars, setting \( m = p = 1 \).

When the system matrices \( A(t) \), \( B(t) \), \( C(t) \), and \( D(t) \) remain constant over time, the continuous-time linear dynamical system simplifies to a linear time-invariant (LTI) system, represented in \cref{eq:2}. This LTI system can be transformed into a discrete-time linear dynamical system, defined by \cref{eq:3}, using discretization techniques. A common method in the SSM literature is zero-order hold (ZOH) discretization \cite{franklin2002feedback}, shown in \cref{eq:4}.

\begin{equation} \label{eq:2}
    \begin{gathered}
        h'(t) = Ah(t) + Bu(t) \\
        y(t) = Ch(t) + Du(t)
    \end{gathered}
\end{equation}

\begin{equation} \label{eq:3}
    \begin{gathered}
        h_t = \bar{A}h_{t-1} + \bar{B}u_t \\
        y_t = Ch_t + Du_t
    \end{gathered}
\end{equation}
\begin{equation}\label{eq:4}
    \begin{gathered}
        \bar{A} = \exp(\Delta A) \\
        \bar{B} = (\Delta A)^{-1}\exp(\Delta A -  I)\Delta B
    \end{gathered}
\end{equation}

\subsection{Selective State Space Model (Mamba)}

Mamba \cite{gu2023mamba} extends the discrete-time linear dynamical system by introducing a timescale parameter, \( \Delta \), which transforms the continuous variables \( A \) and \( B \) into their discrete counterparts, \( \bar{A} \) and \( \bar{B} \). Beyond discretization, Mamba relaxes the time-invariance constraint on the system matrices by introducing a \textit{selection} mechanism. This mechanism allows certain parameters—specifically \( \Delta \), \( B \), and \( C \)—to vary over time as functions, \( s \), of the input \( u \). The formulations are defined in \cref{eq:6}.

\begin{equation}\label{eq:6}
    \begin{gathered}
        s_B(u) = Linear_N(u) \\
        s_C(u) = Linear_N(u) \\
        s_{\Delta}(u) = Broadcast_{D}(Linear_{1}(u)) \\
        \Delta = \tau_{\Delta}(Parameter + s_{\Delta}(u))
    \end{gathered}
\end{equation}

The \( Linear_d \) is a parameterized linear projection to dimension \( d \), and \( \tau_{\Delta} = softplus \). Since the selection mechanism loses equivalence to the convolution form in equation (\ref{eq:4}), Mamba further incorporates a work-efficient parallel algorithm, called \textit{associate scan} \cite{10.5555/1280094.1280110}, into its GPU kernel implementation to enable parallel computation of the system.

\subsection{VMamba} 

The original Mamba block was designed for 1-dimensional input and output, making it unsuitable for computer vision tasks that require 2-dimensional processing. To address this limitation, VMamba \cite{liu_vmamba_2024} introduced a new module called 2D-Selective-Scan (SS2D), which adapts Mamba for 2D input and output. The SS2D module consists of three steps: cross-scan, selective scan (Mamba block), and cross-merge. 

In the cross-scan step, the input feature map is unfolded into four 1D sequences, each capturing information from a distinct spatial direction (\cref{fig:vmamba}). These sequences, processed in parallel by the Selective Scan module, encode diverse spatial perspectives critical for 2D feature processing. The cross-merge step then recombines the processed sequences into a 2D feature map, enabling a global receptive field. VMamba stacks multiple SS2D blocks within each layer to construct the complete model.

\section{Experimental Setup}
\label{appendix:experiemnt_setup}


\subsection{Datasets} QuarterMap is evaluated on two standard benchmarks: ImageNet-1K \cite{deng2009imagenet} for image classification and ADE20K \cite{zhou2017scene} for semantic segmentation. For both datasets, only the validation sets are used.

\subsection{Models} Experiments utilize VMamba backbone models \cite{liu_vmamba_2024}, pre-trained on ImageNet-1K. For semantic segmentation, the UperNet framework \cite{xiao2018unified} is used with the VMamba backbone, trained on ADE20K. The VMamba backbone models are available in three configurations: \textit{tiny}, \textit{small}, and \textit{base}, which vary primarily in the number of layers and the dimensions of sequence length $L$ and channel dimension $D$ within the SS2D block.

Each backbone model consists of four layers. In the \textit{tiny} configuration, these layers are arranged as [2, 2, 8, 2], while the \textit{small} and \textit{base} versions use a configuration of [2, 2, 15, 2]. For each configuration, the dimensions $L$ and $D$ are consistent within each layer but vary across layers. Specifically, the channel dimension $D$ doubles, and the sequence length $L$ decreases by a factor of 4 as depth increases.

\subsection{Additional information} The evaluation metric for the image classification task is top-1 accuracy, while for the semantic segmentation task, we utilize all pixel accuracy (aAcc) and mean intersection over union (mIoU) \cite{PASCALVOC2008Everingham} to assess performance. The batch size for image classification is set to 128, and for semantic segmentation, it is limited to 1 due to the variable input sizes in the validation set. All experiments are conducted on a single NVIDIA A100-SXM4 GPU with 40GB of memory.
\section{Additional Experiments}
\label{appendix:additional_exp}

\subsection{Classification Performance on VMamba Varients}
\begin{table*}[ht]
\centering
\small
\caption{Performance comparison of VMamba models with and without QuarterMap (QM) on ImageNet-1K. The results demonstrate that applying QM improves throughput while maintaining comparable accuracy. The additional overhead in ToMe stems from merge and unmerge operations, whereas QM’s overhead arises from the proposed pruning and upsampling stages.}
\begin{tabular}{lccccccc}
\hline
\textbf{Model} & \textbf{Method}       & \textbf{K} & \textbf{Acc@1 (\%)} & \textbf{Throughput} & \textbf{Speedup} & \textbf{Scan Kernel} & \textbf{Add. overheadd} \\
\hline
VMamba-B       & Baseline              & -          & 83.88               & 590            & 1×               & 1.4ms	  & - \\
               & ToMe                  & 3          & 83.52 (-0.36)       & 424            & 0.72×            & \multirow{2}{*}{0.7ms} & \multirow{2}{*}{9.7ms}\\
               & ToMe                  & 2          & 83.07 (-0.81)       & 389            & 0.66×            &  &\\
               & QM (ours)             & 3          & 83.02 (-0.86)       & 654            & 1.11×            & \multirow{2}{*}{0.6ms} & \multirow{2}{*}{0.2ms}\\
               & QM (ours)             & 2          & 82.58 (-1.30)       & 682            & 1.16×            &  &\\
\hline
VMamba-S       & Baseline              & -          & 83.64               & 811            & 1×               & 1.1ms & -\\
               & ToMe                  & 3          & 83.09 (-0.55)       & 575            & 0.71x            & \multirow{2}{*}{0.5ms} & \multirow{2}{*}{7.3ms}\\
               & ToMe                  & 2          & 82.79 (-0.85)       & 534            & 0.66x            & &\\
               & QM (ours)             & 3          & 82.42 (-1.22)       & 890            & 1.10×            & \multirow{2}{*}{0.5ms} & \multirow{2}{*}{0.2ms}\\
               & QM (ours)             & 2          & 81.56 (-2.08)       & 921            & 1.14×            & &\\
\hline
VMamba-T       & Baseline              & -          & 82.60               & 1548           & 1×               & 0.5ms & -\\
               & ToMe                  & 3          & 82.15 (-0.45)       & 1218           & 0.79x            & \multirow{2}{*}{0.3ms} & \multirow{2}{*}{3.8ms}\\
               & ToMe                  & 2          & 81.64 (-0.96)       & 1138           & 0.74x            & &\\
               & QM (ours)             & 3          & 81.50 (-1.10)       & 1628           & 1.05×            & \multirow{2}{*}{0.2ms} & \multirow{2}{*}{0.1ms}\\
               & QM (ours)             & 2          & 80.01 (-2.59)       & 1671           & 1.08×            & &\\
\hline
\end{tabular}
\label{tab:quartermap_classification_variants}
\end{table*}

We provide full classification results for VMamba \textit{Tiny}, \textit{Small}, and \textit{Base} on ImageNet-1K with and without QuarterMap. As shown in \cref{tab:quartermap_classification_variants}, applying QuarterMap with \(k=3\) leads to modest accuracy drops of 1.1\%, 1.22\%, and 0.86\% for the Tiny, Small, and Base models, respectively. These losses are offset by consistent throughput improvements, with the \textit{Base} model achieving a \(1.11\times\) speedup, and up to \(1.16\times\) when using \(k=2\). The main gains stem from reduced sequence lengths passed through the Mamba block, directly lowering the computational cost of SSM recurrence and selective attention.

Although VMamba-S achieves slightly higher accuracy and throughput than VMamba-B with QuarterMap, this comparison does not account for the common deployment pipeline in real-world applications. In practice, larger models are often chosen for their superior generalization and transferability, especially in downstream tasks. QuarterMap is designed to fit this workflow by enabling efficient post-training compression of high-capacity models, without retraining or access to the original training data, thus preserving transfer performance while meeting run-time constraints.

This post-training strategy is particularly relevant in scenarios where retraining is costly or infeasible, such as on-device deployment, privacy-sensitive settings, or edge environments. QuarterMap supports this use case by offering a lightweight, architecture-aware pruning method that requires no model reconfiguration or finetuning, making it well suited for efficient deployment of pre-trained VMamba models.

\subsection{Semantic Segmentation on ADE20K}
\begin{table}[ht]
\centering
\caption{Performance comparison of VMamba-Upernet models with baseline and QuarterMap (QM) methods with \(k=3\) on ADE20K semantic segmentation.}
\begin{tabular}{lccc}
\hline
\textbf{Backbone} & \textbf{Method}       & \textbf{Acc (\%)}   & \textbf{mIoU (\%)}      \\
\hline
Base & Baseline              & 83.7                & 50.96                  \\
                 & QM (ours)             & 82.94 (-0.76)       & 49.21 (-1.75)          \\
\hline
Small & Baseline              & 83.47               & 50.6                   \\
                 & QM (ours)             & 82.5 (-0.97)        & 48.81 (-1.79)          \\
\hline
Tiny & Baseline              & 82.44               & 47.93                  \\
                 & QM (ours)             & 81.22 (-1.22)       & 44.99 (-2.94)          \\
\hline
\end{tabular}
\label{tab:quartermap_segmentation_perf}
\end{table}
We evaluate QuarterMap on semantic segmentation using VMamba-Upernet \cite{liu_vmamba_2024, xiao2018unified} backbones on ADE20K \cite{zhou2017scene}. As shown in \cref{tab:quartermap_segmentation_perf}, QuarterMap incurs only a 0.76\% decrease in all-pixel accuracy (aAcc) and a 1.75\% drop in mean Intersection over Union (mIoU) in \textit{Base} model. For the \textit{Small} and \textit{Tiny} variants, aAcc decreases by 0.97\% and 1.22\%, and mIoU by 1.79\% and 2.94\%, respectively.

\subsection{BloodMNIST Class-Wise Accuracy}
\begin{table}[h]
\centering
\small
\caption{Class-wise accuracy on BloodMNIST before and after applying QuarterMap. QuarterMap preserves fine-grained classification performance across all classes.}
\rowcolors{2}{gray!10}{white}
\begin{tabular}{lcc}
\toprule
\textbf{Class} & \textbf{Baseline Acc. (\%)} & \textbf{QM Acc. (\%)} \\
\midrule
Basophil              & 97.95 & 97.95 \\
Eosinophil            & 99.68 & 99.68 \\
Erythroblast          & 96.78 & 96.78 \\
Immature granulocytes & 96.55 & 96.72 \\
Lymphocyte            & 99.59 & 99.59 \\
Monocyte              & 94.72 & 94.72 \\
Neutrophil            & 96.25 & 96.40 \\
Platelet              & 100.00 & 100.00 \\
\bottomrule
\end{tabular}
\label{tab:quartermap_bloodmnist_classwise}
\end{table}

To further evaluate QuarterMap's stability in fine-grained settings, we report class-wise performance on BloodMNIST in MedMNIST \cite{medmnistv1, medmnistv2}. As shown in \cref{tab:quartermap_bloodmnist_classwise}, accuracy remains nearly identical across all categories, confirming that QuarterMap preserves semantic precision in medical imaging tasks. This finding demonstrate that QuarterMap generalizes effectively to VMamba-derived architectures and can be confidently applied in domain-specific deployments where accuracy preservation is critical.

\subsection{Extended Restuls on Other Architectures}
\begin{table*}[ht]
\centering
\small
\caption{Accuracy comparison of QuarterMap on CNN, Transformer, and Mamba models for ImageNet-1K classification. QM significantly impacts CNNs, causes a larger accuracy drop in Transformers, and is specifically designed to work on VMamba models.}
\rowcolors{2}{gray!10}{white}
\begin{tabular}{lccc}
\toprule
\textbf{Model} & \textbf{Type} & \textbf{Baseline Acc. (\%)} & \textbf{QM (Ours) Acc. (\%)} \\
\midrule
ConvNeXt-B     & Conv           & 84.89 & 45.71 \\
ConvNeXt-T     & Conv           & 82.94 & 48.26 \\
EfficientNetv2-L & Conv         & 75.75 & 64.71 \\
EfficientNetv2-M & Conv         & 72.92 & 64.05 \\
DeiT-B         & Transformer    & 81.80 & 79.50 \\
DeiT-S         & Transformer    & 79.83 & 74.31 \\
Swin-B         & Transformer    & 85.17 & 82.91 \\
Swin-S         & Transformer    & 83.21 & 81.22 \\
Swin-T         & Transformer    & 81.18 & 78.56 \\
ViM-Base       & Mamba (1D)       & 80.40 & 71.00 \\
ViM-Small      & Mamba (1D)       & 80.50 & 73.20 \\
PlainMamba-L2  & Mamba (4D variant) & 81.60 & 79.30 \\
PlainMamba-L1  & Mamba (4D variant) & 77.70 & 73.30 \\
VMamba-B       & Mamba (4D)       & 83.88 & 83.02 \\
VMamba-S       & Mamba (4D)       & 83.64 & 82.42 \\
VMamba-T       & Mamba (4D)       & 82.60 & 81.50 \\
\bottomrule
\end{tabular}
\label{tab:quartermap_all_archs}
\end{table*}

To complement the analysis in the main text, we provide extended results of applying QuarterMap to additional CNNs, Transformers, and SSM-based architectures. As shown in \cref{tab:quartermap_all_archs}, the variants of each model are reported. The results illustrate QuarterMap’s relative effectiveness across architecture families, with VMamba achieving the most favorable trade-off between accuracy and throughput. In contrast, QuarterMap significantly degrades CNN performance and moderately affects ViTs and 1D-scanning SSMs like ViM and PlainMamba \cite{Yang_2024_BMVC}.

\section{Ablation Study}
\label{appendix:ablation_study}

\begin{figure}
    \centering
    \includegraphics[width=0.45\columnwidth]{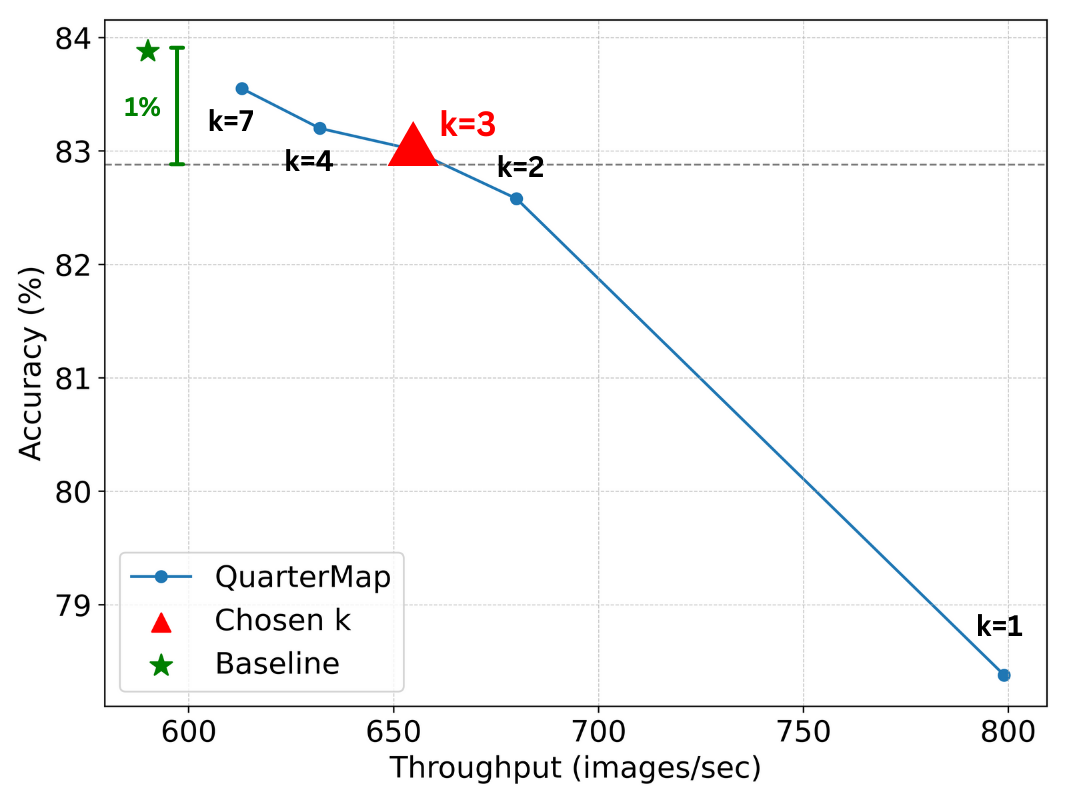}
    \caption{Pareto-optimal analysis from different block selection interval \(k\).}
    \label{fig:figure_pareto_optimal}
\end{figure}

\begin{figure*}[h]
    \centering
    \includegraphics[width=1\textwidth]{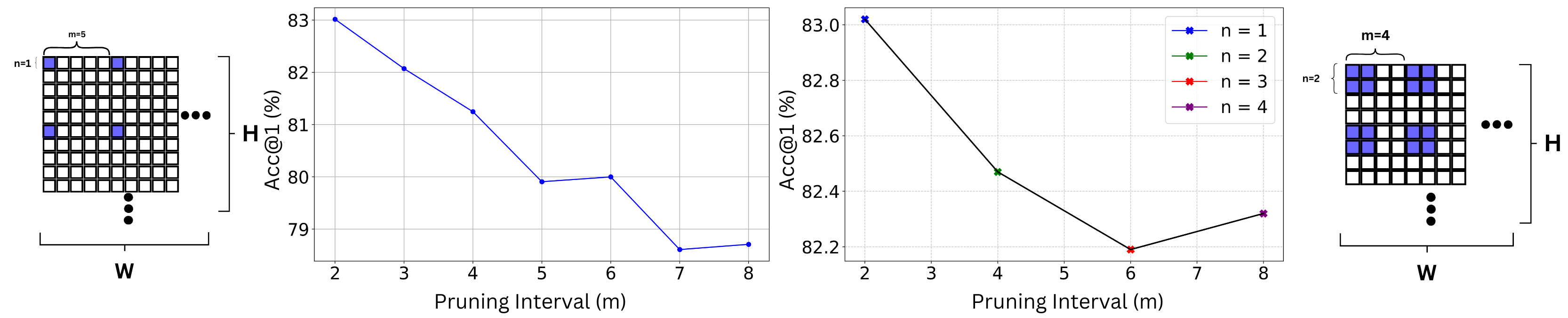}
    \caption{Ablation studies on feature map pruning in QuarterMap on ImageNet-1K classification.}
    \label{fig:ablation_feature_map_pruning}
\end{figure*}

\subsection{Block selection}

We investigate the impact of the block selection interval (\(k\)) for applying QuarterMap to the VMamba-B model on the ImageNet-1K classification task. As shown in \cref{fig:figure_pareto_optimal}, selecting smaller values for 
(\(k\)) yields greater throughput gains but comes at the cost of significant accuracy degradation. To guide this trade-off, we highlight the 1\% accuracy difference in the figure, demonstrating that \(k=3\) represents a reasonable choice based on the Pareto-optimal curve.

Additionally, we assess the impact of applying QuarterMap to different layers of the VMamba-B model, uniformly pruning all blocks within a layer. As shown in \cref{tab:ablation_study_model_layer}, applying QuarterMap to the first layer results in the most significant accuracy drop, highlighting the critical role of early blocks in encoding fundamental low-level features essential for downstream tasks. This finding aligns with prior computer vision studies emphasizing the sensitivity of initial layers to pruning.

In contrast, applying QuarterMap to the third and deepest layer yields the largest throughput improvement due to its higher computational load and greater resilience to pruning. These results underscore the trade-off between accuracy retention and computational efficiency, emphasizing the importance of strategic layer selection when applying QuarterMap.

\subsection{Feature map pruning methods}

We conduct two ablation studies to investigate different methods for pruning feature maps when applying QuarterMap to the ImageNet-1K classification task. These experiments utilize the VMamba-B model, applying QuarterMap with \(k=3\) starting from the second layer. The ablation studies focus on varying the pruning interval \(m\) and and the number of consecutively retained tokens \(n\). 

In the first study, we retain one pixel out of every \(m\) pixels in both spatial dimensions of the feature map (\(n=1)\). For instance, when \( m = 4 \), the original \( H \times W \) feature map is reduced to \( \lceil H/4 \rceil \times \lceil W/4 \rceil \), as in top-left of \cref{fig:ablation_feature_map_pruning}. Results indicate that model accuracy decreases as  \(m\) reflecting the trade-off between pruning granularity and accuracy. Smaller intervals preserve more spatial information, leading to better performance, whereas larger intervals result in greater information loss. Interestingly, comparable accuracy is observed across certain intervals (\textit{e.g.}, \(m=5, 6\) and \(m=7, 8\)) likely due to similar numbers of retained pixels despite differences in interval size.

In the second study, we select \( n \) continuous pixels out of every \( m \) pixels in both spatial dimensions, as depicted in the top-right of \cref{fig:ablation_feature_map_pruning}. The results, shown in the bottom-right of ~\cref{fig:ablation_feature_map_pruning}, indicate that none of the tested configurations outperformed the baseline (blue point). This observation supports our hypothesis that maintaining a critical density of spatial information is crucial for preserving model accuracy.
\begin{table}[ht]
\centering
\caption{Ablation study on applying QuarterMap to different model layers. The throughput is measured in images per second (img/s).}
\begin{tabular}{S[table-format=1]S[table-format=2.2]S[table-format=4]S[table-format=1.2, table-align-text-post=false]}
\toprule
\textbf{Model Layer} & \textbf{Acc@1 (\%)} & \textbf{Throughput} & \textbf{Speedup} \\
\midrule
{Baseline}           & 83.88               & 590                 & 1×               \\
1                    & 78.75               & 645                 & 1.09×            \\
2                    & 83.00               & 618                 & 1.05×            \\
3                    & 80.48               & 748                 & 1.27×            \\
4                    & 83.79               & 597                 & 1.01×            \\
\bottomrule
\end{tabular}
\label{tab:ablation_study_model_layer}
\end{table}

\begin{table}[ht]
\centering
\caption{Comparison of different upsampling methods in terms of accuracy and throughput measured in images per second.}
\begin{tabular}{lcc}
\hline
\textbf{Method} & \textbf{Acc@1 (\%)} & \textbf{Throughput} \\
\hline
Baseline   & 83.88 & 590 \\
Nearest    & 83.02 & 654 \\
Bilinear   & 82.97 & 205 \\
Bicubic    & 82.04 & 161 \\
\hline
\end{tabular}
\label{tab:upsample_method}
\end{table}

\subsection{Upsampling}

We assess the impact of different upsampling methods, specifically nearest neighbor, bilinear, and bicubic, on accuracy (Acc@1) and throughput, utilizing the VMamba-B model on the ImageNet-1K classification task. As shown in \cref{tab:upsample_method}, both bilinear and bicubic upsampling lead to a notable reduction in throughput. This is attributed to their computationally intensive interpolation processes, which involve calculations across multiple neighboring pixels, thereby imposing a substantial computational burden. In contrast, nearest neighbor upsampling achieves higher throughput through simpler calculations, making it a more efficient choice for implementing QuarterMap.


\end{document}